\documentclass[authoryear,preprint,12pt]{elsarticle}

\usepackage{soul}
\usepackage{amssymb}
\usepackage{lipsum}
\usepackage[colorlinks=true,citecolor=blue]{hyperref}
\usepackage{lineno}
\usepackage[dvipsnames]{xcolor}
\usepackage{graphicx}
\usepackage{subcaption}
\usepackage{amsmath}
\usepackage{algorithm}
\usepackage{algorithmic}
\usepackage{array}
\usepackage{amsthm}
\usepackage{subcaption}
\usepackage{multicol}
\usepackage{tikz}
\usetikzlibrary{arrows.meta, positioning, shapes.geometric}
\usepackage{xltabular}
\newcolumntype{Z}[1]{>{\hsize=#1\hsize}X}
\usepackage{threeparttable} 
\usepackage{booktabs}
\usepackage{multirow}
\usepackage{ragged2e}
\usepackage{comment}
\usepackage[top=2.5cm, bottom=2.5cm, left=2.5cm, right=2.5cm]{geometry}
\newtheorem{definition}{Definition}
\newtheorem*{problem}{Problem}

\makeatletter
\def\ps@pprintTitle{%
	\let\@oddhead\@empty
	\let\@evenhead\@empty
	\def\@oddfoot{}%
	\let\@evenfoot\@oddfoot}
\makeatother

\begin{document}

\begin{frontmatter}

	\title{Group Effect Enhanced Generative Adversarial Imitation Learning for Individual Travel Behavior Modeling under Incentives}

        \author[rvt]{Yuanyuan Wu}
        \ead{yuanwu@kth.se}

        \author[rvt]{Zhenlin Qin}
        \ead{zhenlinq@kth.se}

        \author[rvt2]{Leizhen Wang}
	\ead{leizhen.wang@monash.edu}

        \author[rvt3]{Xiaolei Ma}
        \ead{xiaolei@buaa.edu.cn}

        \author[rvt]{Zhenliang Ma\corref{cor1}}
        \ead{zhema@kth.se}


    \begin{abstract}
        Understanding and modeling individual travel behavior responses is crucial for urban mobility regulation and policy evaluation. The Markov decision process (MDP) provides a structured framework for dynamic travel behavior modeling at the individual level. However, solving an MDP in this context is highly data-intensive and faces challenges of data quantity, spatial-temporal coverage, and situational diversity. To address these, we propose a group-effect-enhanced generative adversarial imitation learning (gcGAIL) model that improves the individual behavior modeling efficiency by leveraging shared behavioral patterns among passenger groups. We validate the gcGAIL model using a public transport fare-discount case study and compare against state-of-the-art benchmarks, including adversarial inverse reinforcement learning (AIRL), baseline GAIL, and conditional GAIL. Experimental results demonstrate that gcGAIL outperforms these methods in learning individual travel behavior responses to incentives over time in terms of accuracy, generalization, and pattern demonstration efficiency. Notably, gcGAIL is robust to spatial variation, data sparsity, and behavioral diversity, maintaining strong performance even with partial expert demonstrations and underrepresented passenger groups. The gcGAIL model predicts the individual behavior response at any time, providing the basis for personalized incentives to induce sustainable behavior changes (better timing of incentive injections).
    \end{abstract}


\end{frontmatter}

\section{Introduction}\label{introduction}

    The ever-evolving urban mobility landscape, particularly in the post-pandemic context, presents ongoing challenges for city planners, policymakers, regulators, and operators. A deeper understanding of travelers' behavioral responses is essential for effectively managing and adapting to dynamic urban mobility demand.
    Such Travel Behavior Response (TBR) refers to how individuals adjust their travel choices and decisions in reaction to interventions such as pricing policies, infrastructure changes, demand management strategies, or service interruptions. 
    TBR modeling has been conducted from both short-term and long-term perspectives. The short-term analysis examines the stable behavioral changes at a certain time after an intervention \citep{anupriya2018impact,henn2011surveying}, while the long-term analysis focuses on the temporal modeling of the behavior change process \citep{zhao2018detecting,ma2020behavioral}. 
    
    Traditional travel behavior modeling has primarily relied on statistical methods, machine learning (ML) techniques, and deep learning (DL) approaches to analyze and predict travel choices. However, these approaches struggle to fully capture the dynamic, sequential, and policy-sensitive nature of individual travel responsive decisions. Statistical models, such as discrete choice models (DCM) \citep{ben1985discrete,ben2002hybrid}, provide interpretable frameworks for decision-making but often assume static preferences and limited sequential dependencies over time. ML techniques, including clustering and supervised learning, enhance pattern recognition and prediction accuracy but lack an explicit decision-making framework for adaptive behavior \citep{wang2021comparing,sun2023routine}. DL methods leverage neural networks, such as recurrent neural networks (RNNs) and transformers \citep{feng2018deepmove,cui2018travel}, to improve temporal sequence modeling but remain largely black-box models with limited policy interpretability. To address these limitations, we model the TBR problem as an Markov Decision Process (MDP), which provides a principled framework by formulating travel behavior as a sequential decision-making problem under choice uncertainties. 
    
    MDPs explicitly incorporate state transitions, action choices, and reward structures, allowing for the integration of imitation learning (IL) techniques to infer latent preferences and underlying decision-making strategies.
    IL is a powerful data-driven Artificial Intelligence (AI) technique \citep{osa2018algorithmic} and has emerged as a useful approach for analyzing human choice strategies based on trajectory data from sources such as smart cards, GPS, and cellular calls \citep{zhao2023deep,song2024state}. It relies on a dataset of expert trajectories $\tau={(s_t,a_t)}$, where $s_t$ represents the states at time $t$, and $a_t$ is the corresponding action taken by the expert. The goal of IL is to learn a policy $\pi(a|s)$ that mimics the expert’s behavior as closely as possible. 
    
    Generally, IL methods are categorized into three types. The basic approach is called Behavior Cloning (BC) \citep{torabi2018behavioral}. In BC, the agent learns a direct mapping from states to expert actions using standard classification or regression techniques as a supervised learning problem. It is simple, but its effectiveness highly depends on the quality of the expert data \citep{codevilla2019exploring}.
    
    The second is Inverse Reinforcement learning (IRL), which aims to extract the value of the choice $a$ people make in a given state $s$, i.e., reward function $R(s,a)$ \citep{ng2000algorithms}. An RL algorithm can then use this learned reward function to derive the desired policy $\pi(a|s)$. IRL is particularly useful when the goal is to recover specific reward functions of interest. It has been widely applied in various domains, including identifying food deliveryman's route preferences \citep{liu2020integrating}, personalized route recommendation \citep{liu2022personalized}, taxi drivers' routing behaviors \citep{zhao2023deep}, and valuing travelers' activity-based choices \citep{song2024state}. However, the two-step nature of IRL, first inferring the reward function and then deriving the optimal policy, makes it computationally intensive. Moreover, it typically demands high-quality expert demonstrations to achieve a reliable performance.
    
    The other approach, Generative Adversarial Imitation Learning (GAIL) \citep{goodfellow2014generative}, offers the advantages of bypassing intermediate IRL steps and directly learns a policy from the observed data. Inspired by Generative Adversarial Networks (GANs) \citep{goodfellow2014generative}, GAIL employs the adversarial training approach with two neural networks: Generator/Policy ($G$) and Discriminator ($D$). The task of $G$ is to learn and generate behaviors/actions that the discriminator cannot differentiate from the expert’s, while $D$ distinguishes between the behaviors generated by $G$ and the true behaviors by experts. When $D$ cannot distinguish behaviors generated by $G$ from the true ones, then $G$ has successfully imitated the true behavior. 
    Such adversarial training encourages policies that generalize well to expert behavior, reducing the risk of overfitting to specific trajectories. 
    To also benefit from adversarial training, researchers extended this idea to inverse reinforcement learning, leading to advanced methods such as Adversarial Inverse Reinforcement Learning (AIRL) \citep{fu2017learning}, which serves as a benchmark in this study. 
    
    Most GAIL-based transport studies focus on generating vehicle trajectories \citep{choi2021trajgail}, modeling driving behavior \citep{kuefler2017imitating,bhattacharyya2022modeling}, learning taxi drivers' passenger seeking strategies \citep{zhang2019unveiling,zhang2020cgail,pan2020xgail}, using real vehicle trajectory data. This study leverages the GAIL architecture to learn the human mobility dynamics from discrete travel trajectory data, such as public transport fare card records that capture activity times and locations.
    However, the effectiveness of directly applying GAIL to an individual's observed trajectories can be limited by data sparsity, as personal travel data often spans only a narrow range of contexts, trip purposes, and transport modes. How to model and predict behavior in unseen or less frequently observed situations is of particular interest. While aggregating data across the population may seem like a natural solution, it does not necessarily ensure sufficient diversity or coverage for effective and generalizable policy learning.

    To address this challenge, we propose a group effect conditioned GAIL framework (gcGAIL). It builds on the principles introduced in Conditional Generative Adversarial Networks (cGANs) \citep{mirza2014conditional}. cGANs extend the standard GAN framework by incorporating conditional variables to guide the generation process. Unlike traditional GANs, where the generator produces outputs based solely on random noise, cGANs introduce additional side information (e.g., class labels) into both the generator and discriminator to guide and direct the learning process. cGANs has been used to generate qualitative attributes, such as trip purpose when labeled trip chain demonstrations are limited \citep{kim2022imputing}.
    The proposed gcGAIL categorizes passengers into groups based on shared travel characteristics. These group features/labels are used to augment each passenger's trajectory data by incorporating data from other passengers within the same group, thereby enhancing the model's ability to learn individual behavior more effectively. By leveraging behavioral similarities among travelers, this approach enriches individual-level learning with broader contextual information. The effectiveness of the proposed method is validated through comparisons with benchmark models and an ablation study.

    The main contributions of this study are as follows: 1) Propose the gcGAIL method to model individual-level longitudinal behavioral responses under policy incentives; 2) Develop a novel group effect enrichment to mitigate data sparsity and enhance the model generalizability; 3) Conduct extensive experiments and validate the proposed approach using smart card data, demonstrating its effectiveness in reproducing travel behavior over time and space with high demonstration efficiency.  

    The remainder of this paper is structured as follows: Section \ref{sec:methodology} introduces the problem formulation, key challenges, and proposed solution framework for TBR modeling. It also details the inverse TBR learning approach and the gcGAIL training process. Section \ref{sec:experiments} outlines the experimental setup and presents the results. Section \ref{sec:conclusions} concludes the study and discusses limitations and future directions.

\section{Methodology}
\label{sec:methodology}
\subsection{Problem formulation}\label{mdp}
    To model individual TBR under policy interventions, we formulate the problem as an MDP. The MDP is defined by a tuple $\langle \mathcal{S},\mathcal{A},f,R,\gamma \rangle$ as follows:
    \begin{itemize}
        \item \textbf{State space ($\mathcal{S}$)} represents the traveler's current context at time step $t$. It consists of three parts: 1) trip-related features, such as trip origin, trip destination, departure time, travel mode; 2) historical behavior, such as past travel choices, accumulated incentives; 3) external conditions, such as fare policies, network conditions (e.g., congestion level, service availability).
        \item \textbf{Action space ($\mathcal{A}$)} defines the possible travel choices available to the individual, such as: Mode choice (metro, bus, walking, biking, etc.), departure time choice (peak vs. off-peak) and route choice (direct vs. transfer-heavy options).
        \item \textbf{Transition function ($f$)} governs the probability distribution over the next state $s_{t+1}$ given the current state $s_t$ and action $a_t$. It can capture behavioral inertia (e.g., habitual patterns), stochastic external influences (e.g., service disruptions, unexpected delays) and policy-induced adaptations (e.g., fare incentives shifting departure times).
        \item \textbf{Reward function ($R$)} encodes the perceived utility of a travel choice. It can be used to consider monetary cost (fare paid), time cost (travel and waiting time), convenience factors (transfer penalty, congestion discomfort), policy incentives (fare discounts, loyalty programs), etc.
        \item \textbf{Discount factor ($\gamma$)} determines the importance of future rewards, reflecting habit formation and long-term adaptations to policies.
    \end{itemize}
    
    The goal of an MDP is to learn a policy $\pi: \mathcal{S}\rightarrow \mathcal{A}$ that models how individuals adapt their travel behavior over time in response to policy interventions (such as fare discounts). This can be achieved through IRL or GAIL to infer the latent reward function and decision-making strategy from observed trip trajectories that implies travelers' behavior \citep{ng2000algorithms,ziebart2008maximum}. We formally define the key components as follows:

    \begin{definition}[Trajectory $\tau$]
        A trajectory, \(\tau:\{(s_0,a_0),(s_1,a_1),(s_2,a_2),\cdots,(s_t,a_t)\}\), is a sequence of states $s_t$, actions $a_t$, and optionally rewards, generated by an passenger $u$ interacting with an environment. 
    \end{definition}

    \begin{definition}[Policy function $\pi$]
        A policy function, \(\pi(a|s,u)\), specifies the probability distribution of action the passenger $u$ will choose when in state \(s\).
    \end{definition}
    
    \begin{definition}[Reward function $R$]
        A reward function, \(R(s,a)\), reflects the reward received from taking action \(a\) in state \(s\).
    \end{definition}

    With the key components defined, we formally state our problem as follows:
    \begin{problem}[Individual travel behavior learning from trajectory data]
        Given a set of trajectories $\mathcal{T}={\tau_u}$ collected from a set of passengers $\mathcal{U}$, our goal is to inversely learn a policy function $\pi(a|s,u)$ for an individual passenger $u\in\mathcal{U}$, effectively modeling the adaptive travel behavior in response to policy interventions. 
    \end{problem}
    
    Learning individual responsive behavior from trajectory data is challenging in two key aspects. 1) a passenger's reward and policy functions are inherently spatiotemporally dependent, making it difficult to recover these functions for scenarios not explicitly covered in the observed data; 2) passengers exhibit diverse personal characteristics and commuting contexts, resulting in inherently unique reward and policy functions. Developing a distinct model for each individual is impractical and inapplicable. Therefore, the challenge lies in designing a generalized model capable of capturing the decision-making dynamics across diverse individuals.

    We propose a three-stage methodology framework. \textit{Stage 1 - data preparation}: building on existing literature, we extract features from the original smart card dataset that capture various aspects of passengers' travel characteristics to ensure the possibility of individual-level modeling; \textit{Stage 2 - data-driven TBR modeling:} using the appropriately extracted features, we formulate the individual TBR problem as sequential MDP model and construct an RL environment in a data-driven manner; \textit{Stage 3 - conditional TBR learning}: to account for the diversity in passenger behavior, we develop a customized gcGAIL to learn passengers' policy functions effectively.
    
\subsection{Inverse travel behavior response learning}   
\label{sec:solution}        
    Given expert demonstrations \(\tau:(s_0,a_0),(s_1,a_1),(s_2,a_2),\cdots,(s_T,a_T)\) that arise from an optimal policy ($\pi^*$, $\pi_E$) under some unknown reward function ($R(s,a)$), the problem then becomes a typical IRL task \citep{ng2000algorithms}: can we recover the reward function \(R(s,a)\) from this trajectory (Eq.~\ref{eq:irl})? 
    \begin{equation}\label{eq:irl}
        \mathbb{E}[\sum_{t=0}^T\gamma^tR(s_t,a_t)|\pi^*]\geq \mathbb{E}[\sum_{t=0}^T\gamma^tR(s_t,a_t)|\pi] \quad \forall\pi
    \end{equation}
    
    Once we have inferred \(R\), we can use it to derive policies \((\pi\in\Pi)\) of the traveler by solving the Bellman Optimality Equation \citep{sutton1998reinforcement}. This two-step process can be inefficient because solving for the reward function may not directly imply the policy. And after the reward is learned, a policy needs to be found using additional RL steps, which increases computational overhead.
    GAIL offers a more effective way to solve the problem, it bypasses the reward inference step by directly learning a policy through adversarial training. Once trained, it can automatically excel in flexible, non-linear behavior replications.
    
    Introduced by \cite{ho2016generative}, GAIL leverages the principles of GANs \citep{goodfellow2014generative} to directly learn policies that imitate expert demonstrations without the need to explicitly infer the reward function, which is a key difference from traditional IRL. There are two key components of GAIL: Generator (Policy Learner \(\pi_{\theta},G(a|s)\)) and Discriminator \(D(s,a)\). The policy acts as a generator that produces actions based on the current state. Similar to GANs, the Generator tries to `fool' the discriminator by generating actions that resemble expert behavior. The policy is trained to maximize the probability that the discriminator classifies its actions as coming from the expert. While the discriminator tries to distinguish between actions taken by the expert and those taken by the policy \(\pi_{\theta}\). It assigns high probabilities to expert-like actions and low probabilities to non-expert-like actions. When $D$ cannot distinguish actions generated by $G$ from the demonstrations, then $G$ has successfully imitated the true policy. $G$ and $D$ are normally trained simultaneously as if they are playing the \textit{min-max} game with the objective:
    \begin{equation}\label{eq_1}
        \min_{\pi_\theta} \max_{D_\psi} \ \mathbb{E}_{(s,a) \sim \pi_E} \left[ \log D_\psi(s, a) \right] 
        + \mathbb{E}_{(s,a) \sim \pi_\theta} \left[ \log (1 - D_\psi(s, a)) \right] 
    \end{equation}
    where $\pi_E$ is the expert policy (demonstrations), $\pi_{\theta}$ is the policy to be learned (the generator), $D_\psi(s, a)$ outputs the probability that a state-action pair came from the expert. 
    It has been proven to be solvable by strategically choosing a proper convex loss function regularizer, please refer to \cite{ho2016generative}.

    However, the real challenge in our problem lies in the fact that we typically observe only partial demonstrations rather than the full optimal policy.
    Such challenges are prevalent across most data-driven approaches to travel behavior modeling. Directly fitting an individual travel behavior estimation function may fail to model complete traveler's decision-making process. 
     
\subsection{Group effect enrichment}\label{sub:group}
    Providing additional information ($\mathbf{c}$) to GAIL offers a strategic approach for guiding the learning process, addressing challenges related to data sparsity. For example, \cite{zhang2020cgail} applied conditional GAIL (cGAIL) to learn individual taxi driver's operational strategies using vehicle trajectory data. In their study, each \textit{state-action} pair was augmented with condition features, such as the distance from the current location to the driver’s home, the time deviation from the driver’s average work start and end times, and the historical frequency of visits to the current state. They hypothesized that such an augmentation would make the obtained conditional trajectories general. However, their implementation of cGAIL did not consistently outperform standard GAIL (without condition features), as the added information sometimes introduced misleading patterns. 

    To address this, we refine the conditioning approach by grouping condition features to enhance training stability. Recognizing that travelers classified by specific criteria often exhibit similar behavioral patterns, we introduce a group effect enrichment mechanism into cGAIL (gcGAIL). This involves categorizing each trajectory entry of trajectory $\tau_u$ into multiple groups $\mid\mathbf{g}\mid$. 
    The specific grouping methods could be tailored to the research objectives and domain expertise. Taking public transport smart card data as an example, user $u$ can be divided into different groups according to travel characteristics. Each group will imply a general mobility pattern that passenger $u$ may have in a certain aspect. Even if passenger $u$ has limited data, the group-based representation enriches the available information by incorporating insights from similar passengers, enabling a more demonstration efficient approach for individual TBR modeling. 
    
\subsection{gcGAIL framework}
    
\begin{figure}[ht]
        \centering
        \includegraphics[width=0.8\linewidth]{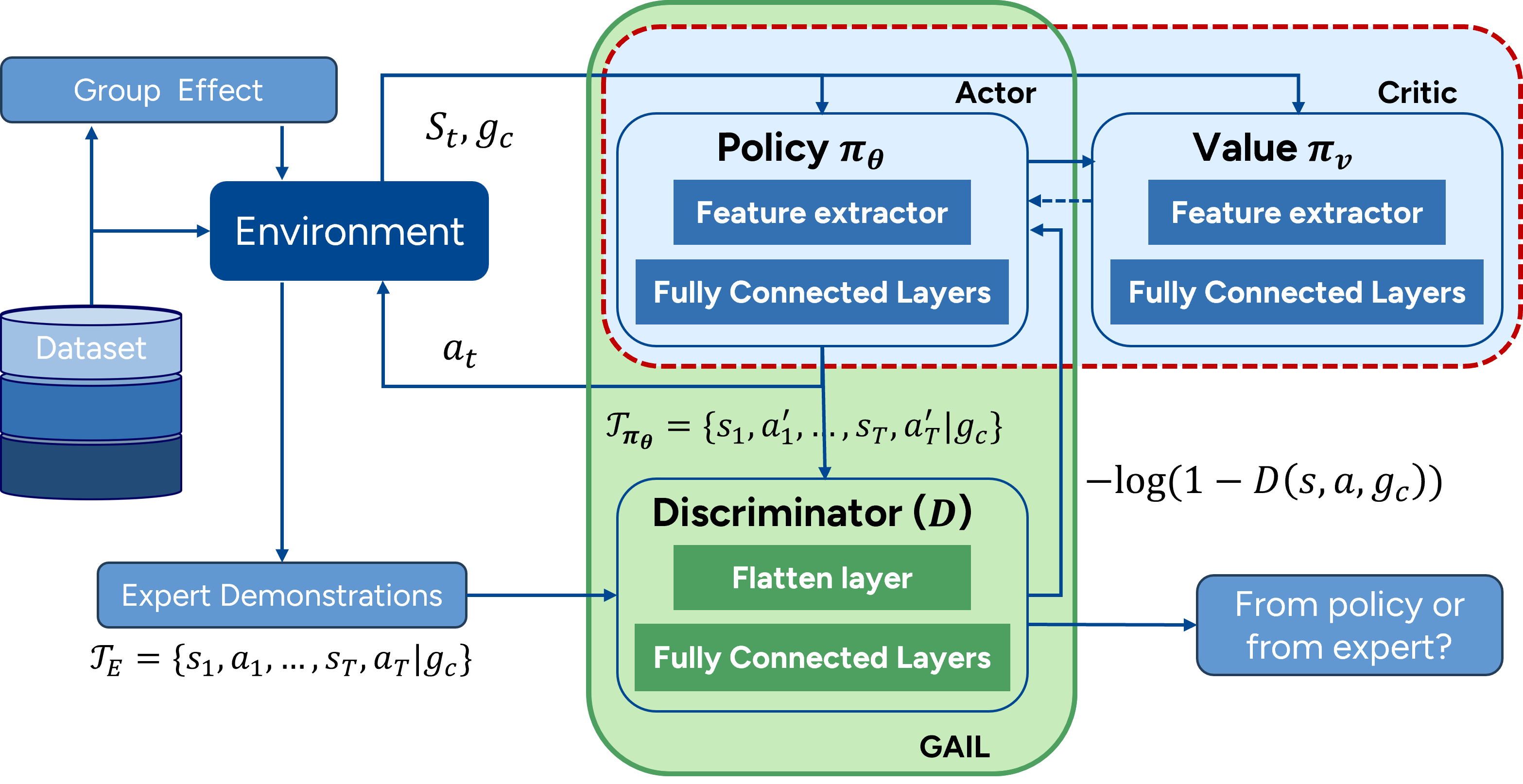}
        \caption{gcGAIL model for individual travel behavior learning. The policy network acts as the Generator for GAIL. The discriminator is trained with experts and generates policy trajectories, conditioned on $g_c$. The policy is updated using PPO with rewards derived from the discriminator.}
        \label{fig:gcgan}
    \end{figure}

    Figure~\ref{fig:gcgan} shows the gcGAIL model structure. Our proposed gcGAIL contains three different neural networks: policy neural network ($\pi_\theta$, actor), value neural network ($\pi_v$, critic), and discriminator network ($D$, discriminator). The actor (policy) consists of a three-layer network. The hidden layers comprise two dense layers serve as the feature extractor through the \texttt{ReLU} activation function. The last is the output layer with a \texttt{Softmax} activation function, and the output dimension corresponds to the dimension of actions. Critic has similar network structure to that of the actor. The only difference is that the last output layer is linear, and corresponding to the state value $V(s_t)$. Discriminator ($D$), known as the \textit{Reward Neural Network}, has the same hidden layers as the actor's, while the last layer is activated by \texttt{Sigmoid} and outputs a specific probability value as a discounted reward. 

    Instead of using a standard policy gradient method, Proximal Policy Optimization (PPO, \citep{schulman2017proximal}) is applied to stabilize and enhance GAIL policy updates. 
    The process alternates between: collecting trajectories using the current policy $\pi_\theta(a | s, g_c)$; training the discriminator with expert and policy trajectories, conditioned on $g_c$; updating the policy using PPO with rewards derived from the discriminator.
    
    The discriminator $D(s, a, g_c)$ is trained to distinguish between expert and policy trajectories. The objective function for the discriminator is:
    \begin{equation}\label{eq:Dobject}
        \mathcal{L}_D(\psi) = \mathbb{E}_{\tau_E} \left[\log D_\psi(s, a, g_c)\right] + \mathbb{E}_{\tau_{\pi_\theta}} \left[\log(1 - D_\psi(s, a, g_c))\right]
    \end{equation}
    where:
        $\tau_E$ are the trajectories sampled from the expert dataset, paired with the condition $g_c$.
        $\tau_{\pi_\theta}$ are the trajectories sampled from the policy under the same condition $g_c$.
    
    The discriminator outputs a reward signal for the policy and a loss signal for its own optimization.
    The policy $\pi_\theta(a | s, g_c)$ receives rewards from the discriminator:
    \begin{equation}\label{eq:reward}
        r(s, a, g_c) = -\log(1 - D(s, a, g_c))  
    \end{equation}
    
   The difference between the value function and the discriminator output reward is applied as the advantage function to guide the training of the policy network.  \textit{Advantage function} ($\hat{A}_t$), is computed using Generalized Advantage Estimation (GAE) :
    \begin{equation}\label{eq:advantage}
    \hat{A}_t^{\text{GAE}(\gamma, \lambda)} = \sum_{l=0}^{\infty} (\gamma \lambda)^l \delta_{t+l}
    \quad \text{where} \quad
    \delta_t = r_t + \gamma V(s_{t+1},g_c) - V(s_t,g_c)
    \end{equation}
    here $r_t$ is the reward provided by the discriminator, $V(s_{t}, g_c)$ is the value estimate (critic), $\gamma$ is the discount factor, and $\lambda$ is the GAE smoothing parameter. 
   
   The PPO objective is then used to optimize the policy:
    \begin{equation}\label{eq:PPOobjective}
        \mathcal{L}_{\text{PPO}}(\theta) = \mathbb{E}_t \left[ \min \left( \omega_t(\theta) \hat{A}_t, \text{clip}(\omega_t(\theta), 1 - \epsilon, 1 + \epsilon) \hat{A}_t \right) \right]
    \end{equation}
    where:
    $\epsilon$ is the clipping parameter to prevent large updates. $\omega_t(\theta)$ is the probability ratio of the new policy over the old policy, conditioned on $g_c$:
    \begin{equation}
        \omega_t(\theta) = \frac{\pi_\theta(a_t | s_t, g_c)}{\pi_{\theta_\text{old}}(a_t | s_t, g_c)}
    \end{equation}
    
    Algorithm \ref{alg:gcGAIL} illustrates the iterative training process of the gcGAIL model. 
    \begin{algorithm}
        \caption{gcGAIL training process}
        \label{alg:gcGAIL}
        \begin{algorithmic}[1]
            \REQUIRE Expert trajectories $\tau_E$, initial policy parameters $\theta$, initial discriminator parameters $\psi$, PPO hyperparameters: $\gamma$, $\lambda$, $\epsilon$
            \STATE Initialize policy network $\pi_\theta$, discriminator $D_\psi$
            \WHILE{stopping conditions not satisfied}
                \STATE Sample trajectories $\tau_\pi = \{(s, a, s')\}$ using policy $\pi_\theta$ in the environment
                \STATE Compute rewards for each state-action pair with Eq.~\ref{eq:reward}
                \STATE Update $\psi$ using gradient ascent on $\mathcal{L}_D(\psi)$ with Eq.~\ref{eq:Dobject}
                \STATE Compute advantage estimates for $\tau_\pi$ with Eq.~\ref{eq:advantage}
                \STATE Update $\theta$ using gradient ascent on $\mathcal{L}_{\text{PPO}}(\theta)$ with Eq.~\ref{eq:PPOobjective}
            \ENDWHILE
            \RETURN Learned policy $\pi_\theta$
        \end{algorithmic}
    \end{algorithm}
    
\section{Case study}
\label{sec:experiments}
    To evaluate the effectiveness of the proposed gcGAIL model, we conduct a case study on passengers' behavioral responses to a real-world public transport fare discount program. This allows us to examine how gcGAIL captures implicit decision-making strategies of individuals and improves the modeling of travel behavior adaptations under fare-based incentives.
    
\subsection{Data and model settings}\label{sub:featureextraction}
    
\begin{figure}[htbp]
	\centering
	\includegraphics[width=0.7\textwidth]{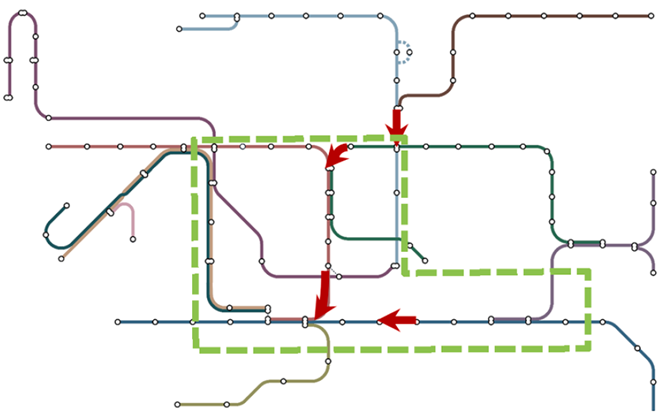}
	\caption{Network, critical links (red arrows), and eligible promotion stations (dashed area) (adopted from \cite{ma2020behavioral}).}
	\label{fig:network_critical_links}
    \end{figure}

    To encourage off-peak travel, Hong Kong Mass Transit Railway (MTR) launched the `early-bird' promotion in September 2014 \citep{MTRHongKong}: \textit{start your day earlier and enjoy a 25\% fare discount!} Available Monday to Friday (except public holidays), from 7:15 to 8:15a.m., passengers using an Adult Octopus can enjoy a 25\% fare discount when exiting any of designated core urban stations (29 stations, see Figure~\ref{fig:network_critical_links}). 
    Nearly all MTR trips are paid for using the Octopus Card. This stored value system deducts fare amounts based on travel distance, and during the promotion period, no fare cap is applied for adult card users \citep{halvorsen2020demand}. 

    The study utilizes Octopus card transaction data, which records key trip details such as anonymized card IDs, card type, tap-in and tap-out times/stations, and fare deductions. Given that the program was launched in September 2014, Automated fare collection (AFC) trip records from July 2014 to October 2015, from 21,347 passengers, are identified and used for the before-and-after analysis. The anonymized ID in the dataset is unique to each passenger across multiple days, allowing for the tracking of individual behavior changes before and after the promotion's implementation. This enables the construction of panel data for modeling purposes.    
    We model passengers' responsiveness to system incentives over time as an MDP, the state features and actions must be carefully defined to align with the MDP framework while ensuring they satisfy its fundamental properties.
    Various factors influences the decisions of passengers, and we extract and summarize into two categories (see Table~\ref{tab:features}): \textit{state features} $s$, which characterizes various statistics of the state from the historical data; \textit{group effect features} $g$, which captures the features of the passenger’s travel characteristics, to augment the individual data for stable and adequate training.
    \begin{table}[!htb]
	\centering
	\caption{Features, definitions, and proxies.}\label{tab:features}
	\small
	\begin{tabularx}{1.0\linewidth}{>{\raggedleft\arraybackslash}
		Z{.1}Z{.9}} \toprule 
        \textbf{Features} & \textbf{Definitions and Proxies} \\
        \midrule 
         \multicolumn{2}{l}{State features}\\
        $l_{home}$ & Inferred home location from the station where most of a passenger's first trips start and last trips end  \\
        $l_{work}$ & Inferred work location from the station where most of a passenger's first trips end and last trips start \\
        $d_t$ & Average tap-in time for morning trips in the most recent period\\
        $e_t$ & Average tap-out time for morning trips in the most recent period\\
        $c_t$ & Average deducted fares for morning trips in the most recent period  \\
        $b_t$ & Average minimum time required to adjust travel to benefit from the promotion in the most recent period\\
        $m_t^p$& Indicating the number of months since the promotion began (with negative values denoting the period before the promotion's start)\\
        $m^s_t$& Average monetary savings gained through the incentive\\
        $w_u$& Whether the inferred work location of passenger $u$ is in specified discounted stations\\
        $\lambda_t$& Latest travel behavior, 1: off-peak travel, 0: otherwise\\
        $\lambda_{t-1}$& Previous month's travel behavior, 1: off-peak travel, 0: otherwise\\\midrule
        \multicolumn{2}{l}{Group effect features, based on data two months before the promotion launch}\\
        $flex_u$& The first trip tap-in time standard deviations on weekdays, representing schedule flexibility. Categorized into quartiles, representing low, low-medium, medium-high, and high flexibility levels\\
        $con_u$& The expected minimum time passenger $u$ has to change to be eligible for the incentive, higher values higher inconvenience; Categorized into quartiles\\
        $dis_u$& Average trip time of morning trips on workdays, indicating the trip distance; Categorized into quartiles\\        
        \bottomrule
	\end{tabularx}
    \end{table}

    \textit{State features} $\mathcal{S}$. Passenger $u$ considers a lot of features associated with their current state $s_t$ to make a decision. Corresponding to the definition in subsection \ref{mdp}, the trip-related features are represented using home location ($l_{home}$), work location ($l_{work}$), the average trip start time ($d_t$), trip end time ($e_t$), trip fare ($c_t$), and minimum shift time ($b_t$). 
    Historical behavior features are crucial for capturing the dynamics of choice behavior over time. We have: `LastMonthMode' ($\lambda_t$), indicating their latest travel behavior, `PreMonthMode' ($\lambda_{t-1}$), indicating the previous month's travel behavior. For the travel behavior for a specific month, we define that a passenger's monthly travel mode is labeled 1 (off-peak) if their off-peak trips exceed half their peak trips in that month; otherwise, it's labeled 0 (peak).

    External conditions in our case study is about the incentive related features, they are represented by: `Months\_Passed' ($m_t^p$), indicating the number of months since the promotion began (with negative values denoting the period before the promotion's start);  `MeanMorningPeakMoneySave' ($m^s_t$), reflecting the average monetary savings gained through the incentive; and `WorkLocIsDiscSta' ($w_u$), indicating whether the inferred work location of passenger $u$ is in specified discounted stations.

    \textit{Group effect features} $\mathcal{G}$.
    Group effect information is essential for our problem, three most influential features shaping responsive behavior identified in previous studies \citep{wang2023data,wu2025data} are selected: `flexibility' ($flex_u$), `inconvenience' ($con_u$) and `travel distance' ($dis_u$). `flexibility' is the schedule flexibility, represented by the standard deviations of the first trip tap-in time on weekdays, higher values indicate higher flexibility. `inconvenience' is the expected minimum time passenger $u$ has to change to be eligible for the incentive, higher values higher inconvenience. Given the absence of sociodemographic and choice information in AFC data, these passenger characteristics were calculated based on each user's Octopus card activity during the two months prior to the promotion policy implementation (July and August 2014 in the case study) \citep{wang2023data}. Their numerical values are used as the \textit{condition features}. To facilitate conditional training, they are further categorized into four different groups (\{1,2,3,4\}) separately using quartiles, which serve as training labels.

    \textit{Actions}  $\mathcal{A}$. In our case study, we focus on whether passengers respond to the incentive by traveling during off-peak hours in a certain month. The actions correspond to departure time choices and are represented as a binary variable: off-peak travel behavior (target behavior, action 1) and all other travel behavior (action 0).

    The state transition is deterministic as time series, we need to inversely reveal the reward function and policy function. 
    
\subsection{Experiment scenarios and training}
    Processed dataset is randomly split into training and testing sets with an 80\% to 20\% ratio. We designed different experiments to validate the model from different perspectives. 
    \begin{itemize}
        \item Experiment I: The entire training dataset was used as expert demonstrations, covering all stations, including all 29 designated discounting stations, in the trajectory demonstrations. 

        \item Experiment II: A subset of stations was randomly selected from the training dataset, covering only half of all stations. Among these, 16 out of the 29 discounting stations were included in the demonstrations. This experiment tests the model's ability to generalize when demonstrations are limited to a subset of stations and in quantity. 

        \item Experiment III: Different proportions ($\in [10\%,30\%,50\%,70\%,90\%]$) of the training dataset were randomly selected as expert demonstrations to assess the model's performance under varying amounts of training data. After the random selection, we assessed the spatial coverage. Except for the 10\% scenario, which covered 98\% of the stations, all other scenarios achieved full station coverage.

        \item Experiment IV: Expert demonstrations were limited by selectively excluding different group-level demonstrations to examine the model's ability to learn from incomplete or missing group-level data. We also assessed the spatial coverage for this set of experiments, confirming that all selected training data achieved full station coverage. 
    \end{itemize}
    
    In particular, we found that the passengers who are in medium-high flexibility ($g_{flex_u} = 3$), low-medium inconvenience ($g_{con_u}=2$) and low-medium travel distance ($g_{dis_u}=2$) exhibit higher responsiveness to the incentive \citep{wu2025data}. To demonstrate the generalization capability of the proposed model, we designed four different scenarios in the experiment IV: scenario \textit{`wf3'}, excludes data from passengers with medium-high flexibility; scenario \textit{`wc2'}, excludes data from passengers with low-medium inconvenience; scenario \textit{`wd2'}, excludes data from passengers with low-medium travel distance; scenario \textit{`ges'}, selectively excludes data based on group-level effects, where the training data does not cover medium-high flexibility, low-medium inconvenience and low-medium travel distance. These scenarios allow us to evaluate how well the model generalizes when specific passenger groups are underrepresented or excluded from the training data.
    
    From Experiment I, we compare the learning accuracy against benchmark methods, including BC, AIRL, GAIL, and cGAIL. Condition features are integrated into the demonstrations/trajectories for both BC and AIRL. GAIL represents the baseline model without group effects and condition features, while cGAIL leverages condition features but omits group effects. gcGAIL incorporates both condition features and group-level effects. 
    Furthermore, we evaluate how the policy learning accuracy is influenced by different GAIL-related model parameters across various experimental settings. We assess the model's ability to generalize when the demonstrations are limited in terms of quantity, spatial-temporal coverage, and behavior diversity.

    All the experiments and evaluations are conducted on a workstation with an Intel Xeon w7-3465X (2.50 GHz) processor, four NVIDIA RTX A6000 GPUs, and 512 GB of memory. The implementation is based on Stable-Baselines3 \citep{stable-baselines3}. All three networks (policy, critic, and discriminator) employ hidden layers with 64 units each. Additional training-related parameters are summarized in Table~\ref{tab:parameters}.
    
    \begin{table}[ht]
        \centering
        \caption{Parameters and stopping conditions for training.}
        \label{tab:parameters}
        \begin{tabular}{ccc|cc}
        \hline
             Parameter& Meaning& Value & Parameter& Value \\\hline
             $\epsilon$& clipping range  & 0.2 & random seed & 0 \\
             $\lambda$ & GAE factor & 0.95 & batch size & 256 \\
             $\gamma$& discount factor & 0.95  & optimization epochs per update & 30  \\
             $\eta$ & learning rate & 0.0001 &  &  \\\hline
             Parameter&\multicolumn{3}{c}{Training stopping conditions}&Value\\\hline
             $X$& \multicolumn{3}{c}{stop if no improvement for $X$ evaluations}& 20\\
             $disc_{acc}$ & \multicolumn{3}{c}{stop if discriminator accuracy is close to 0.5} & $0.5\pm 0.05$\\\hline
        \end{tabular}
    \end{table}

\subsection{Evaluation Metrics}
    Since our action set is binary, confusion matrix is applied to evaluate the accuracy. Specifically, the confusion matrix consists of four key components: True Positives (TP), where the model correctly predicts a positive action, off-peak travel behavior in our case; False Negatives (FN), where the model incorrectly predicts a negative action for an actual positive instance; False Positives (FP), where the model incorrectly predicts a positive action for an actual negative instance; and True Negatives (TN), where the model correctly predicts a negative action.
    From this matrix, key evaluation metrics can be derived \citep{qin2024deepags}:
    \begin{itemize}
        \item \textbf{Accuracy}: Measures overall correctness of predictions.
        \[
        \text{Accuracy} = \frac{TP + TN}{TP + TN + FP + FN}
        \]

        \item \textbf{Precision}: Evaluates the proportion of correctly predicted positive actions among all predicted positives.
        \[
        \text{Precision} = \frac{TP}{TP + FP}
        \]

        \item \textbf{Recall (Sensitivity)}: Measures the proportion of correctly predicted positive actions among actual positives.
        \[
        \text{Recall} = \frac{TP}{TP + FN}
        \]

        \item \textbf{F1 Score}: Harmonic mean of precision and recall, balancing both metrics.
        \[
        \text{F1 Score} = 2 \times \frac{\text{Precision} \times \text{Recall}}{\text{Precision} + \text{Recall}}
        \]
    \end{itemize}

\subsection{Baseline methods comparison}
    Table~\ref{tab:benchmarking} (\textit{left half}) presents the accuracy, precision, recall and f1-score comparison between different models. Results are from Experiment I where the entire training dataset was used as the expert demonstrations. 
    gcGAIL achieves the highest prediction accuracy and demonstrates the best overall balance between precision and recall among the other models. While AIRL yields the highest precision, it suffers from poor recall, indicating a conservative prediction strategy that misses many actual positive cases. Both GAIL and cGAIL outperform AIRL on accuracy, recall and F1 score. But they still fall short of the balanced performance achieved by gcGAIL. In contrast, BC, lacking the advantages of adversarial training, fails to perform the task effectively. 
    \begin{table}[h]
        \centering
        \caption{Evaluation metrics comparison between different models. Results from Experiment I.}
        \begin{tabular}{ccccc|cc}
            \hline
            Model & Accuracy & Precision  & Recall  & F1-score & Acc\_non-Adopters$^*$ & Acc\_Adopters$^*$ \\
            \hline
            BC & 0.28 & 0.27 & \textbf{1.00} & 0.42&0.26$\pm$0.005&0.42$\pm$0.073\\
            AIRL  & 0.82 & \textbf{0.94} & 0.36 & 0.52 &0.84$\pm$0.027&0.69$\pm$0.039 \\
            GAIL  & 0.88 & 0.91 & 0.63 & 0.74&0.90$\pm$0.022&0.77$\pm$0.037 \\
            cGAIL & 0.93 & 0.87 & 0.85 & 0.86&0.95$\pm$0.021&0.79$\pm$0.026 \\
            gcGAIL & \textbf{0.95} & 0.92 & 0.91 & \textbf{0.91}&0.97$\pm$0.015&0.86$\pm$0.021\\
            \hline
            \multicolumn{7}{l}{$^*$\textit{prediction accuracy} $\pm$ \textit{standard deviation} for non-adopters and adopters}
        \end{tabular}
        \label{tab:benchmarking}
    \end{table}

\begin{figure}[htbp]
        \centering
        \includegraphics[width=\linewidth]{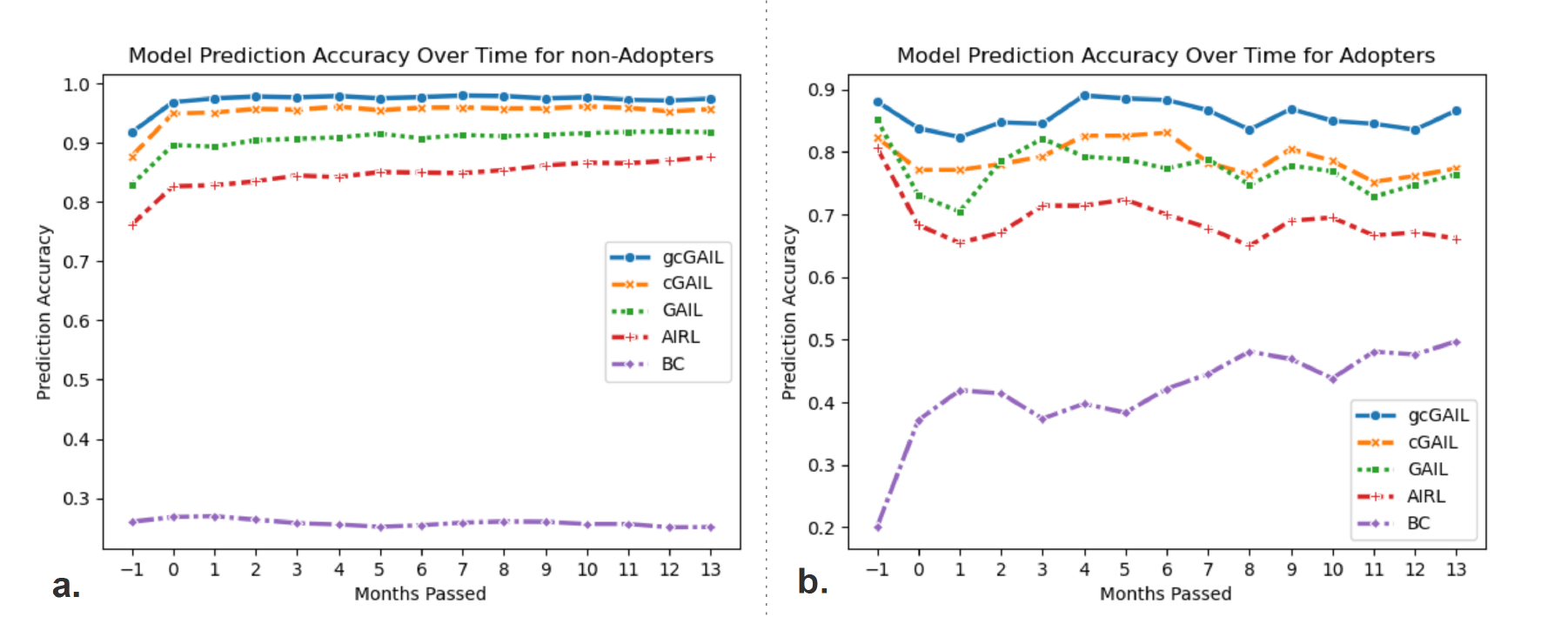}
        \caption{Model prediction accuracy of different models over time of (a.) non-Adopters and (b.) Adopters. Results are from Experiment I. The month passed 0' means September, 2014.}
        \label{fig:heatmap_acc}
    \end{figure}

    Using a time-series change point detection method, passengers are classified as either adopters or non-adopters of the promotion \citep{wang2023data}.
    Figure~\ref{fig:heatmap_acc} illustrates the prediction accuracy of different models over time, separately for non-adopters and adopters. The corresponding mean accuracy and standard deviation are reported in Table~\ref{tab:benchmarking} (\textit{right half}). Our proposed gcGAIL consistently outperforms other methods, achieving higher accuracy with lower standard deviation, thereby demonstrating its effectiveness in capturing dynamic passenger behavior in response to fare incentives.        

    \subsection{Ablation study of conditional and group effects}
\begin{figure}[htbp]
        \centering
        \includegraphics[width=\linewidth]{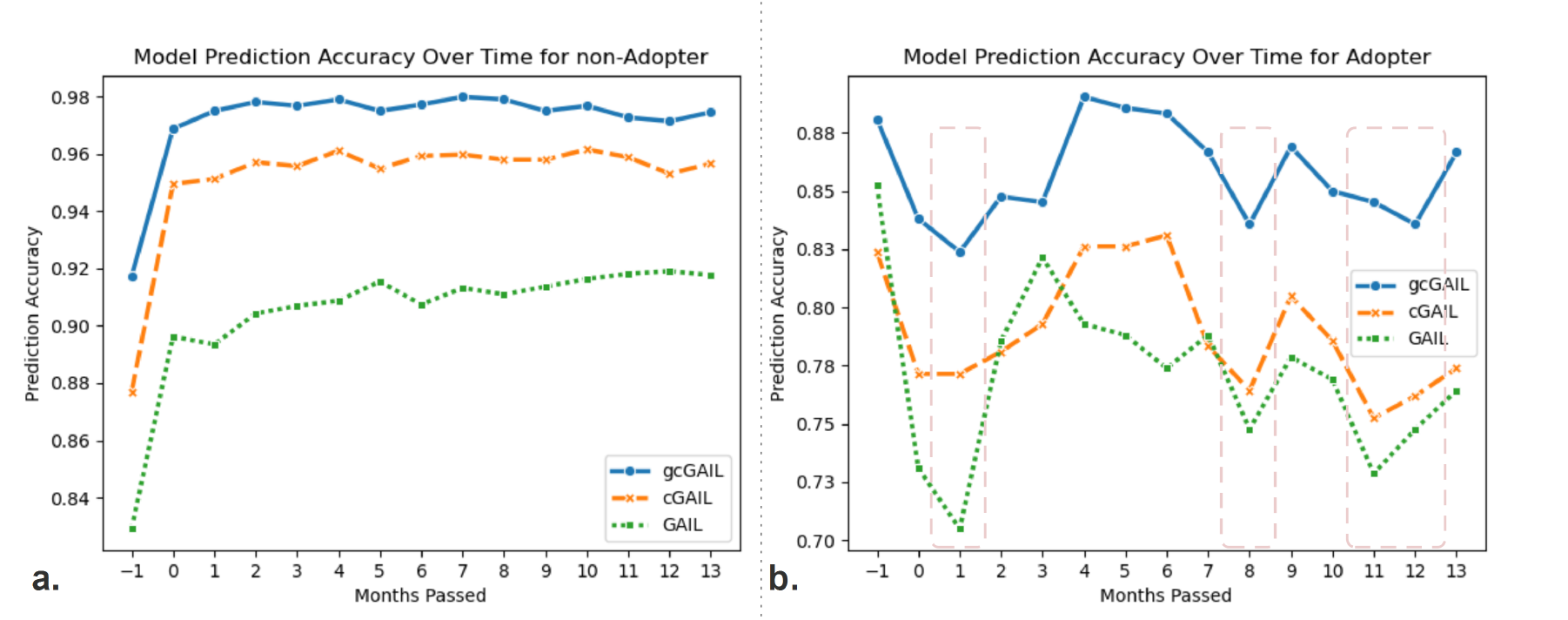}
        \caption{Model prediction accuracy of GAIL-related models over time upon (a.) non-Adopters and (b.) Adopters. Results are from Experiment I.}
        \label{fig:pre_acc}
    \end{figure}

    The comparison between GAIL-based models serves as a structured ablation study to evaluate the contribution of conditional and group-level effects. 
    Figure~\ref{fig:pre_acc} details the prediction accuracy of GAIL-based models for both incentive adopters and non-adopters. All three models perform well for non-adopters, who generally maintain their previous travel patterns after the introduction of the incentive. But all models show slightly lower accuracy for adopters, whose behavior tends to be more dynamic. 
    
    For adopters, the results demonstrate that gcGAIL consistently achieves the highest accuracy (0.86$\pm$0.021) despite some fluctuations, highlighting the benefit of group-level effects in improving predictive performance. GAIL shows lower accuracy and exhibits higher fluctuation, particularly in the early months, suggesting that without conditional features, it struggles to capture the complex and dynamic behavior of adopters. The inclusion of additional contextual information in cGAIL and gcGAIL helps mitigate this early volatility by providing more guided and stable training. However, cGAIL achieves only moderate accuracy (0.79$\pm$0.026), confirming that group-level effects play a critical role in capturing complex passenger behavior dynamics.

\begin{figure}[ht]
        \centering
        \includegraphics[width=\linewidth]{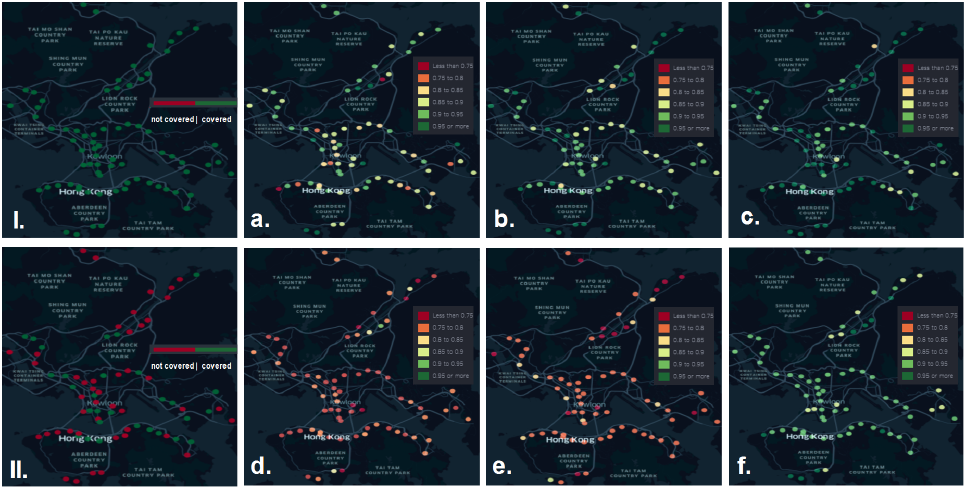}
        \caption{Model prediction accuracy of GAIL, cGAIL, gcGAIL upon stations in the first and second experiments. I. shows the station coverage in Experiment I, with a., b., and c. presenting the results from GAIL, cGAIL, and gcGAIL, respectively; II. shows the station coverage in Experiment II, with d., e., and f. presenting the results from GAIL, cGAIL, and gcGAIL, respectively.}
        \label{fig:acc_loc}
    \end{figure}

    In addition, Figure~\ref{fig:acc_loc} compares the prediction accuracy of GAIL, cGAIL and gcGAIL across stations from Experiments I and II. Notably, in Experiment II, only half of the stations are included in the expert demonstration data, making it a more challenging generalization task. Despite this, gcGAIL demonstrates a superior ability to reproduce travel patterns spatially, maintaining higher accuracy across various locations. This further validates the effectiveness of incorporating group-level effects in improving predictive performance both temporally and spatially.

\subsection{Model generalization}
    The group effect enrichment is proposed to address potential data sparsity, which is introduced in three aspects in Experiments II, III, and IV: limited spatial-temporal coverage, quantity, and passenger diversity. 
    Under limited spatial-temporal coverage and less demonstration size, see Figure~\ref{fig:acc_loc}, the overall accuracy of all models declines. However, gcGAIL still demonstrates relatively strong performance. This suggests that gcGAIL is less sensitive to spatial variations in the demonstration data, further highlighting its effectiveness in capturing behavioral variations. A possible reason for this robustness is that the group effect features do not explicitly encode location-specific information, allowing the model to generalize better across different stations.

\begin{figure}[ht]
        \centering
        \includegraphics[width=\linewidth]{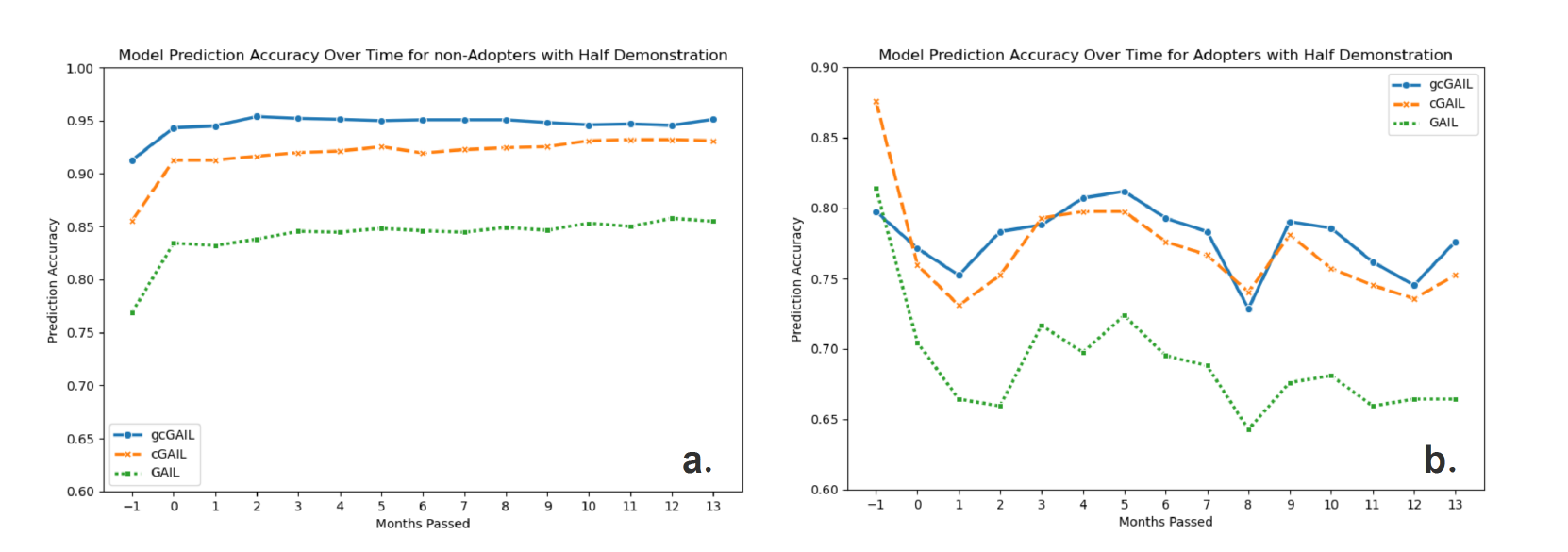}
        \caption{Model prediction accuracy of GAIL, cGAIL, gcGAIL over time of (a.) non-Adopters and (b.) Adopters with half demonstrations. Results are from Experiment II.}
        \label{fig:acc_sec}
    \end{figure}

    The overall prediction accuracy of GAIL, cGAIL, gcGAIL in Experiment II is $0.82$, $0.89$, and $0.92$, representing drops of approximately $6.8\%$, $4.3\%$, and $3.2\%$ compared to Experiment I. Figure~\ref{fig:acc_sec} illustrates their prediction accuracy over time for both non-adopters and adopters in \textbf{Experiment II}. For non-adopters, gcGAIL effectively captures patterns and generalizes well, with the highest performance throughout the observation period. However, for adopters, the performance of all models declines and exhibits noticeable fluctuations at specific time points (e.g., 1, 8, and 11–12 month(s) passed). These fluctuations suggest potential shifts in passenger behavior during these periods, which will be analyzed in detail later.

\begin{figure}[ht]
        \centering
        \includegraphics[width=\linewidth]{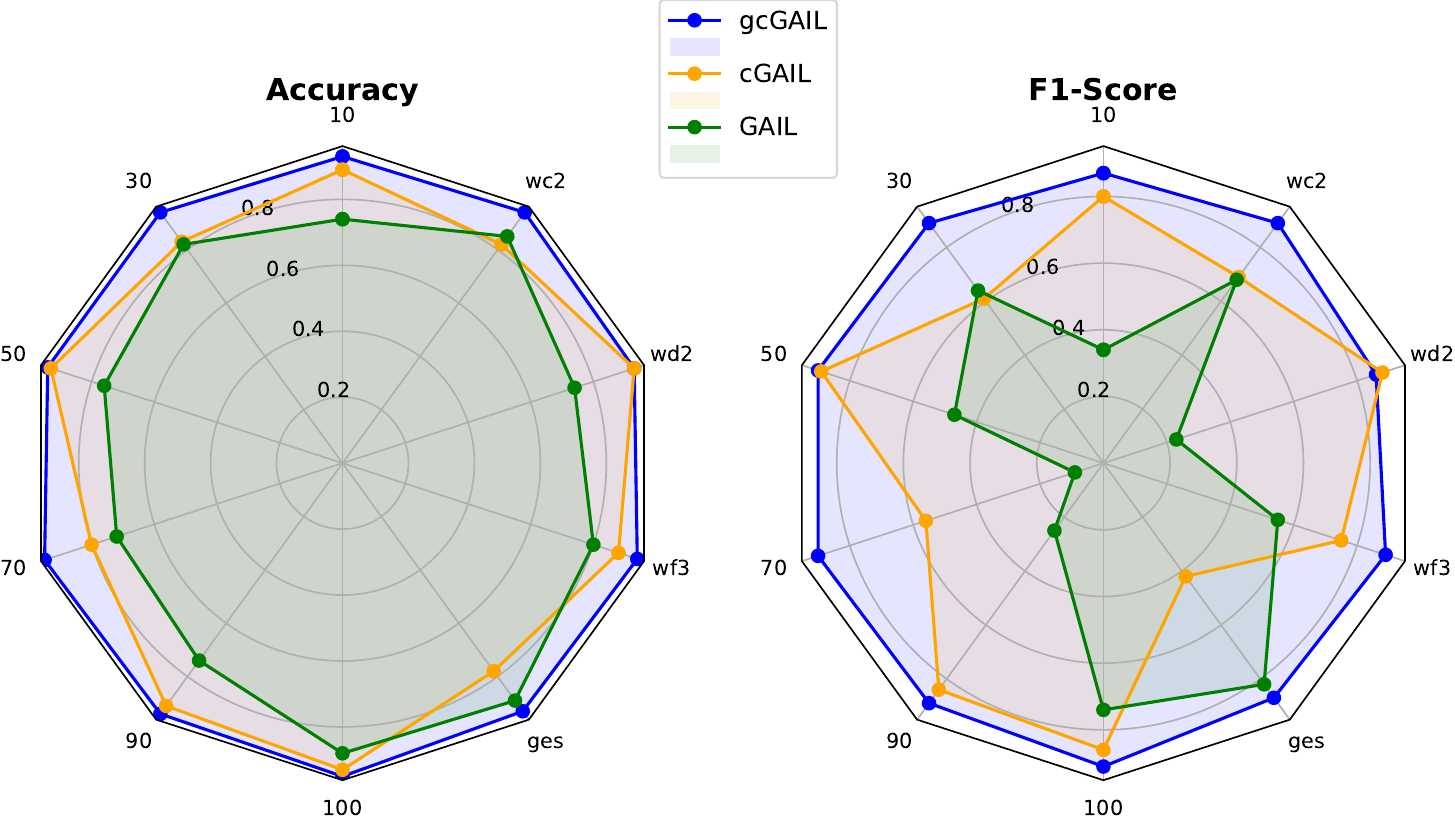}
        \caption{Model prediction accuracy and F1-Score of GAIL, cGAIL, gcGAIL from different scenarios. \textit{`10-100'} represent the proportions of training dataset was used in the training; \textit{`wf3'} scenario, training data does not cover the passengers in medium-high flexibility group; \textit{`wc2'} scenario, training data does not cover the passengers in low-medium inconvenience group; \textit{`wd2'} scenario, training data does not cover the passengers in low-medium travel distance group; \textit{`ges'} means the group-effect is selected and that training data does not cover medium-high flexibility group, low-medium convenience group and low-medium travel distance group.}
        \label{fig:acc_scenarios}
    \end{figure}

    Experiment III is designed to evaluate the model's generalization capability under varying amounts of expert demonstrations. Experiment IV, on the other hand, assesses how well the models generalize when specific passenger groups are underrepresented or excluded from the training demonstrations. 
    Figure~\ref{fig:acc_scenarios} presents the results from both experiments. 
    From the radar chart, the performance of gcGAIL remains stable, showing consistent accuracy and F1-scores regardless of the demonstration quantity and passenger diversity. In contrast, the other two models exhibit inconsistent performance: prediction accuracy does not consistently improve with demonstration size, and cGAIL does not consistently outperform GAIL. Interestingly, we observed an unexpected decline in accuracy and F1-score for GAIL and cGAIL when the demonstration size increased to 70\%. While the underlying cause remains unclear, this suggests that the randomness in the training data may have influenced the model's performance, whereas gcGAIL appears to be more robust to such variability.

    The \textit{`ges'} scenario in Experiment IV represents a special case of limited demonstrations, where the most responsive trajectories are excluded. This makes the remaining demonstrations less diverse compared to other scenarios. The reduced behavioral diversity benefits the training of baseline GAIL, leading to higher accuracy and F1-scores than cGAIL under this scenario (Table~\ref{tab:ges}). 
    Since cGAIL relies on condition features to guide training, the absence of key demonstrations from the most responsive behaviors makes it difficult for cGAIL to generalize to unseen behaviors.
    In contrast, gcGAIL generalizes well even without the most responsive demonstrations. By specifying behavior similarity at the group level, gcGAIL can leverage shared behavioral patterns to predict unseen behaviors. For example, even if it does not encounter cases with $g_{flex_u} = 3, g_{con_u} = 2, g_{dis_u}= 2$, it can still generalize by learning from other trajectories.
    
    \begin{table}[ht]
        \centering
        \caption{Evaluation metrics comparison between different models under scenario \textit{`ges'}.}
        \begin{tabular}{ccc|ccc|ccc}
            \hline
            \multicolumn{3}{c|}{Overall Accuracy}&\multicolumn{3}{c|}{F1-score}&\multicolumn{3}{c}{Acc\_Adopters$^*$}\\
            \hline
              GAIL & cGAIL & gcGAIL & GAIL & cGAIL & gcGAIL& GAIL & cGAIL & gcGAIL\\\hline
             0.90 &0.78& 0.93&	0.82 & 0.42 &	0.87 & 0.79 & 0.64& 0.83\\\hline
            \multicolumn{9}{l}{$^*$model prediction accuracy for adopters}
        \end{tabular}
        \label{tab:ges}
    \end{table}
    
\subsection{Behavior response pattern analysis}\label{subsub:rpa}
   
    Adopters were further classified into six distinct types based on changes in their temporal travel patterns before and after the promotion \citep{wang2023data}:
    \begin{itemize}
        \item \textbf{Early adopters}: Passengers who adopt the promotion \textbf{within the first two months} after its implementation.
        \item \textbf{Late adopters}: Passengers who adopt the promotion \textbf{two months after} its introduction.
        \item \textbf{Early morning adopters}: Passengers who shift their travel from the early morning (6:15–7:15 AM) to the off-peak period (7:15–8:15 AM) to receive the discount.
        \item \textbf{Morning peak adopters}: Passengers who shift their travel from the morning peak (8:15–9:15 AM) to the off-peak period (7:15–8:15 AM) to receive the discount.
        \item \textbf{Attrition adopters}: Passengers who initially adopt the promotion but later discontinue its use.
        \item \textbf{Sustained adopters}: Passengers who adopt the promotion at a certain point and continue using it thereafter.
    \end{itemize}
    
\begin{figure}[htbp]
        \centering
        \includegraphics[width=\linewidth]{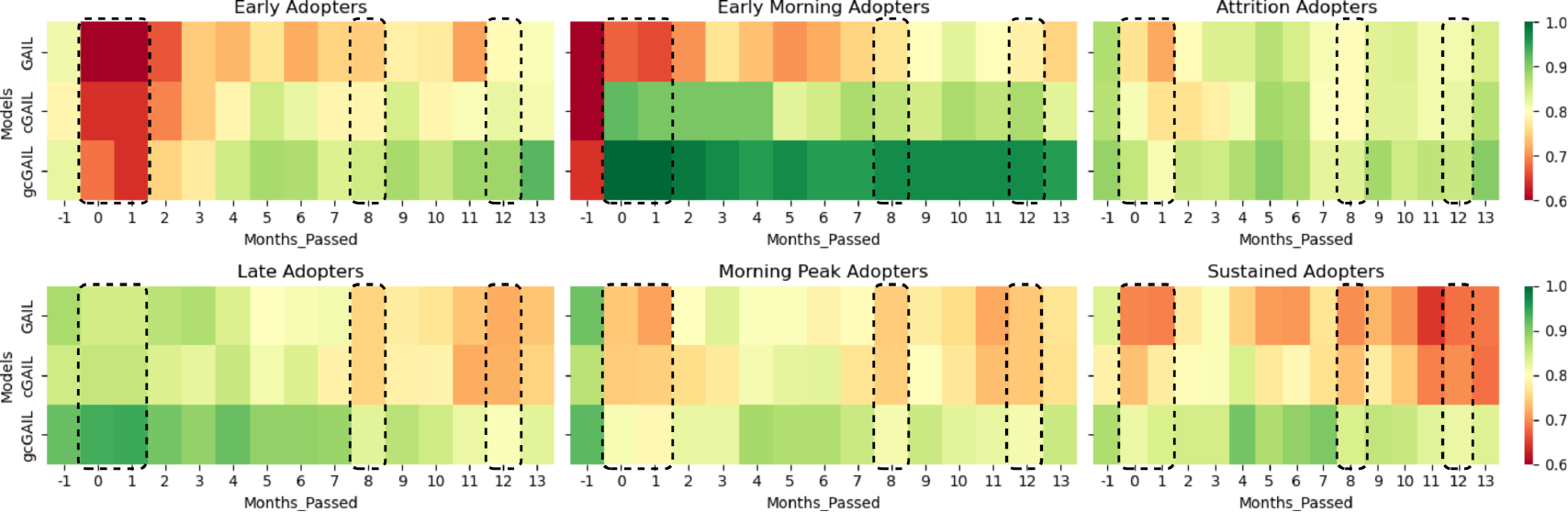}
        \caption{Model prediction accuracy on different types of adopters. Results are from Experiment I.}
        \label{fig:pre_acc_dif}
    \end{figure}

    Figure~\ref{fig:pre_acc_dif} compares the prediction accuracy of GAIL-based models across different adopter types. In all cases, gcGAIL maintains the highest performance. The observed fluctuations basically align with those in Figure~\ref{fig:pre_acc}b, with key turning points occurring during the initial two months, around the eighth month, and approximately one year after the promotion was introduced. These turning points suggest that passenger responses to fare incentives evolve over time, reflecting phases of early experimentation, mid-term disengagement, and longer-term stabilization. The ability of the gcGAIL to maintain high prediction accuracy throughout these phases highlights its robustness in capturing not only spatial but also temporal heterogeneity in travel behavior. From a policy perspective, these time points may represent critical windows for intervention to support long-term adoption and sustained behavior changes (a key objective of demand management).
    
    For early adopters who adopt the promotion within the first two months, all models show lower accuracy during this period, as it represents their adaptation phase to the incentive. The lower accuracy in the first two months for early adopters suggests that passenger behavior during this period is highly dynamic and less predictable, likely due to individual trial-and-error responses to the incentive. And the fact that accuracy improves after this initial phase implies that the models successfully capture stable behavioral patterns once they emerge.
    
    Similarly, for late adopters who adopt the promotion at least two months after its implementation, all models show higher accuracy during the first two months, as these passengers have not yet adjusted their travel behavior in response to the incentive. And the subsequent decrease in accuracy suggests that passengers begin to actively respond to the promotion, leading to greater variability in their travel patterns. 
    
    All models are trained to capture the dynamic nature of passenger responses to fare incentives, with gcGAIL demonstrating particularly strong performance in inferring the behavior of early morning adopters. The consistently higher prediction accuracy for this group suggests that their travel behavior adjustments are more stable and sustainable over time. This observation aligns with prior findings, which showed that 61.2\% of early morning adopters were classified as sustained adopters. 
    Morning peak adopters, who shift their travel from the peak period (8:15–9:15 AM) to the off-peak period (7:15–8:15 AM) to receive the discount, represent the primary target of the incentive program. Before the promotion, all models achieve relatively higher accuracy in predicting their behavior. However, after the promotion’s implementation, prediction accuracy declines, with greater variability across models, indicating that their responses are more volatile and unpredictable. This underscores the challenges in sustaining behavioral change.

    For attrition adopters, all models exhibit only marginal performance. For the more desirable sustained adopters, only gcGAIL demonstrates higher predictive accuracy. While its performance fluctuates slightly over time, these variations reflect the inherent dynamics of responsiveness, suggesting that gcGAIL is able to capture behavioral shifts to a meaningful extent.
    Overall, the proposed gcGAIL consistently outperforms other models in predicting passenger travel behavior under fare incentives, particularly excelling in reproducing long-term travel patterns for sustained and early adopters.
    However, it may struggle with capturing rapid, short-term behavioral shifts, suggesting the need for further improvements to enhance its ability to model transitional and volatile adaptations in response to incentives. 
    
\section{Conclusions}
\label{sec:conclusions}
    Data-driven modeling of individual travel behavior often faces challenges related to data availability and sparsity. To address this, we proposed gcGAIL, a group-effect-enhanced generative adversarial imitation learning model designed to learn travel behavior response from smart card-derived passenger trip trajectories. Using a fare-discount promotion as a case study, we assessed the effectiveness of gcGAIL by comparing it against GAIL, cGAIL, and AIRL in terms of accuracy, generalization, and pattern demonstration efficiency.

    Our findings demonstrate that gcGAIL consistently outperforms other models in capturing passenger travel dynamics, particularly in reproducing long-term behavioral adaptations. Its ability to infer travel patterns over both time and space highlights its effectiveness in modeling individual-level travel behavior responses. Notably, gcGAIL exhibits lower sensitivity to spatial variation, even when expert demonstrations provide only partial spatial coverage. Moreover, gcGAIL maintains strong performance under scenarios of data sparsity and behavioral diversity (experiments with underrepresented or excluded passenger groups). While GAIL and cGAIL experienced a notable decline in prediction accuracy, gcGAIL maintained satisfactory prediction accuracy despite limited data availability. 
    
    These results highlight the efficiency of gcGAIL in reproducing individual behavior choices over time. However, the precise number of expert demonstrations required to train a satisfactory model remains an open question and is beyond the scope of this study. Additionally, group effects are represented using quartile-based categories for the interpretation. It may not fully reflect the complexity and diversity of passenger responses. More refined or context-specific groupings could be explored in future work to align with different research objectives. 

\section*{Acknowledgements}
    We would like to acknowledge KTH Digital Futures (cAIMBER and CIML4MOB projects) and TRENoP research centers in Sweden for their funding supports. 
    
    During the preparation of this work the authors used ChatGPT 4o in order to check the English grammar. After using this tool, the authors reviewed and edited the content as needed and take full responsibility for the content of the publication.

\bibliographystyle{elsarticle-harv}
\bibliography{tbrmodelling}

\end{document}